\pdfoutput=1

\documentclass[11pt]{article}

\usepackage[preprint]{coling}
\usepackage[utf8]{inputenc}

\usepackage{times}
\usepackage{latexsym}

\usepackage[T1]{fontenc}

\usepackage[utf8]{inputenc}

\usepackage{microtype}

\usepackage{inconsolata}

\usepackage{graphicx}
\usepackage{subfigure}
\usepackage{booktabs}
\usepackage{amsmath}
\usepackage{amsthm}
\usepackage{enumitem}
\usepackage{color,amsfonts}
\usepackage{multirow,array}
\usepackage{tikz}
\usepackage{pgfplots}
\usepackage{amssymb,mathtools}
\usepackage[linesnumbered,ruled,vlined]{algorithm2e}
\usepackage{cleveref}
\usepackage{caption}

\usepackage{tabularx}
\usepackage{pifont}
\usepackage{cancel}
\usepackage{lipsum}
\usepackage{fvextra}
\usepackage{booktabs}
\usepackage{multirow}
\usepackage{xcolor}
\usepackage{ragged2e}
\usepackage{makecell}

\newcommand{\model}[1]{\textsc{#1}\xspace}
\newcommand{\llama}{\model{Llama-3.1}}
\newcommand{\llamasmall}{\model{Llama-3.1 8B}}
\newcommand{\llamalarge}{\model{Llama-3.1 70B}}
\newcommand{\alma}{\model{Alma-R 13B}}
\newcommand{\qwen}{\model{QWen2}}
\newcommand{\mistral}{\model{Mistral}}
\newcommand{\gptfouro}{\model{gpt-4o mini}}
\newcommand{\gemma}{\model{Gemma2 9B}}
\newcommand{\nllb}{\model{NLLB}}
\newcommand{\nllbsmall}{\model{NLLB-600M}}
\newcommand{\nllbmedium}{\model{NLLB-1.3B}}
\newcommand{\nllbbig}{\model{NLLB-3.3B}}
\title{Proverbs Run in Pairs: Evaluating Proverb Translation Capability of Large Language Model}

\author{Minghan Wang, Viet-Thanh Pham, Farhad Moghimifar, Thuy-Trang Vu \\
  Department of Data Science \& AI, Monash University \\ 
  \texttt{\{firstname.lastname\}@monash.edu}
}

\begin{document}
\maketitle
\begin{abstract}
Despite achieving remarkable performance, machine translation (MT) research remains underexplored in terms of translating cultural elements in languages, such as idioms, proverbs, and colloquial expressions.
This paper investigates the capability of state-of-the-art neural machine translation (NMT) and large language models (LLMs) in translating proverbs, which are deeply rooted in cultural contexts.
We construct a translation dataset of standalone proverbs and proverbs in conversation for four language pairs.
Our experiments show that the studied models can achieve good translation between languages with similar cultural backgrounds, and LLMs generally outperform NMT models in proverb translation.
Furthermore, we find that current automatic evaluation metrics such as BLEU, CHRF++ and COMET are inadequate for reliably assessing the quality of proverb translation, highlighting the need for more culturally aware evaluation metrics.\footnote{Dataset and code will be released upon acceptance.}

%

\end{abstract}

\section{Introduction}
Translating multi-word figurative expressions, particularly idioms and proverbs, have long been a challenge in MT due to their meanings diverging from the literal interpretation of individual words~\citep{Constant2017CoLI,zaninello-birch-2020-multiword}. Previous research in neural machine translation (NMT) has primarily focused on translating idiomatic expressions to capture their figurative meanings in the source language and accurately convey them in the target language~\citep{isabelle-etal-2017-challenge, fadaee-etal-2018-examining, avramidis-etal-2019-linguistic}. 
The development of large language models (LLMs) for MT has demonstrated a reduction in overly literal translations, improving the quality of idiomatic translations~\citep{raunak-etal-2023-gpts}.
However, the translation of proverbs, which are short popular sayings conveying cultural beliefs, has received comparatively less attention.

%




Translating proverbs presents additional challenges beyond simply preserving figurative meaning. Proverbs are often deeply rooted in cultural contexts, and their translation requires careful cultural adaptation. This involves translating culture-specific terms, paraphrasing, and considering widely accepted versions of the proverb that resonate in the target language~\citep{newmark2003textbook}.
Recent research by \citet{Liu2024AreMultilingualLLMs} has shown that while LLMs possess some degree of knowledge about proverbs, their capability of reasoning with proverbs, especially in dealing with figurative proverbs, remains limited. In light of these challenges and recent developments, using proverbs as a proxy for cultural common grounds, our study examines the ability of current NMT and LLM-based MT models in proverb translation to better understand how existing MT models handle cross-cultural elements. In particular, we aim to answer the following research questions: (i) \emph{Can current MT methods, particularly LLM-based MT systems, handle proverb translation?}; (ii) \emph{What are the roles of conversation contexts and prompts in proverb translation ability of LLM-based MT?}; and (iii) \emph{Can current automatic evaluation metrics measure the accuracy of translating cultural nuances?}




To address these research questions, we first expand the existing multicultural proverbs and sayings dataset \citep{Liu2024AreMultilingualLLMs} into an English-centric proverb translation dataset.
As proverbs can also appear in conversations, we further mine the Proverb in Conversation (PiC) dataset by extracting proverb usage from movie subtitles. These datasets include five languages representing diverse geographical areas: English, German, Bengali, Indonesian, and Mandarin Chinese. 
We then conduct extensive evaluations of state-of-the-art NMT systems and multiple LLM families on these datasets to assess their proverb translation capabilities.

Our experiments reveal that current MT models demonstrate a certain level of proficiency in handling proverb translations, with notably better performance observed when translating between languages from similar cultural areas. 
Furthermore, we observe that the reliability of existing automatic evaluation metrics is insufficient for accurately assessing the cultural nuance in proverb translation. Specifically, we make the following contributions
\begin{itemize}[noitemsep,topsep=2pt,parsep=2pt,partopsep=2pt]
    \item We construct a proverb translation and proverb in conversation translation dataset in four translation pairs to facilitate the future research in figurative language translation.
    \item Our experiments provide insights into the roles of context and prompting in the performance of proverb translation tasks.  Larger models already learn the meaning of proverbs, hence, adding the proverb interpretation or example does not help. On the other hand, conversation context in dialogue format plays a more important role.
    \item We also find that the current automatic evaluation metrics such as BLEU and COMET as well as the LLM-as-a-judge are unreliable and very sensitive to small lexical changes when evaluating figurative translation quality.
\end{itemize}

\section{Proverb Translation Dataset}
\subsection{Standalone Proverb Translation Dataset} 
Proverbs are fixed expressions that convey traditional wisdom and deeply rooted in lived experiences and socio-cultural contexts. This makes proverbs an excellent lens through which to evaluate how well the current MT model captures and translates cultural information. 
 To facilitate such analysis, we extend the MAPS dataset \citep{Liu2024AreMultilingualLLMs} which is a collection of proverbs from multiple languages, including English (\textsc{En}), German (\textsc{De}), Bengali (\textsc{Bn}), Mandarin Chinese (\textsc{Zh}) and Indonesian (\textsc{Id}).\footnote{We omit Russian (\textsc{Ru}) in our study due to lack annotators.}
 These languages come from diverse geographical regions and exhibit varying linguistic structures and resource availability according to \citet{Joshi2020StateFateLinguistic}.
Each proverb in MAPS dataset is companied with its explanation, the machine translation into English, and a label indication whether the proverb is figurative or literal.
The figurative proverbs have meanings that differ from their literal expressions, while the literal ones convey meaning directly.
This dataset allows us to evaluate not only the translation quality but also how well models can handle figurative language. \Cref{tab:maps-data} provides detailed statistics.



\begin{table}[t]
\begin{center}
\scalebox{0.9}{
  \begin{tabular}{lrrr}
  \toprule
  \textbf{Language} &\multicolumn{1}{c}{\textbf{\#Prov.}} & \multicolumn{1}{c}{\textbf{\#Fig.}} & \multicolumn{1}{c}{\textbf{Region}}  \\
  \midrule
  English (\textsc{En})  & 424 & 232 & Western Europe\\
  Bengali (\textsc{Bn}) & 340 & 272 & South Asia\\
  German (\textsc{De}) & 334 & 183 & Western Europe\\
  Indonesian (\textsc{Id}) & 341 & 267 & Southeast Asia\\
  Chinese (\textsc{Zh}) & 334 &  143 & East Asia\\
  \bottomrule                 
  \end{tabular}
 }
   \vspace{-2mm}
   \caption{MAPS dataset statistics including number of proverbs (\#Prov.), number of figurative proverbs (\#Fig.) and geography region.}
   \label{tab:maps-data}
   \end{center}
   \vspace{-2em}
\end{table}

To ensure the quality of the dataset, we recruit annotators to verify and post-edit the machine-translated proverbs. These annotators were fluent in both English and native speakers of Chinese, Indonesian, or Bengali, with experience in either professional or volunteer translation work. For German-to-English translations, although the annotators were not native German speakers, they were proficient in both German and English. The annotation process was structured to ensure the cultural and linguistic accuracy of the translations. 

Annotators were presented with proverbs in their native language alongside the machine-generated English translations. Their primary task was to evaluate the correctness of these translations and post edit where necessary. Additionally, they are also tasked with identifying context-dependent proverbs whose meanings can shift based on the situation in which they are used.  For English proverbs, the annotators are also asked to provide equivalent proverbs in their native languages that convey similar meanings, when possible. The native proverb provided will serve as a translation reference during the evaluation of from English translation. The details of annotation protocols and addition analysis can be found in \Cref{appx:data}. 

\subsection{Proverb in Conversation Mining}

OpenSubtitles~\citep{Lison2016OpenSubtitles2016ExtractingLarge} is a multilingual parallel corpora of movie and TV subtitles. It contains culturally rich conversation and spans multiple languages, making it a good source to mine the parallel corpus of proverb usage. 
However, as OpenSubtitles is often included in LLM pretraining corpus, we perform data contamination analysis in \S\ref{sec:contam} and find that contamination is not significant enough to bias the evaluation. 

\begin{table}[t]
\centering
\resizebox{0.9\columnwidth}{!}{%
\begin{tabular}{@{}l|rrr|rrr@{}}
\toprule
 & \multicolumn{3}{c|}{\textbf{From-En}} & \multicolumn{3}{c}{\textbf{To-En}} \\
\textbf{Lang} & \textbf{P1} & \textbf{P2} & \textbf{P3} & \textbf{P1} & \textbf{P2} & \textbf{P3} \\ \midrule
\textsc{Bn} & 89 & 69 & 42 & 511 & 76 & 8 \\
\textsc{De} & 7028 & 2000 & 1540 & 1903 & 1755 & 1129 \\
\textsc{Id} & 2459 & 1969 & 1214 & 51 & 44 & 13 \\
\textsc{Zh} & 3456 & 2000 & 272 & 1498 & 1488 & 827 \\ \bottomrule
\end{tabular}%
}
\caption{The statistics of the mined subtitles in each phrase: \textbf{P1: Initial Mining \S\ref{sec:p1_filter}}; \textbf{P2: Fine-grained Filtering \S\ref{sec:p2_filter}}; \textbf{P3: Human Evaluation \S\ref{sec:p3_filter}}. 
}
\label{tab:filter_statistic}
\vspace{-5mm}
\end{table}
\subsubsection{Initial Mining with Source-side Proverb} 
\label{sec:p1_filter} 
We first collect potential translation pairs which containing proverbs from OpenSubtitles dataset. In this step, we preprocess both proverbs and translation pairs with lemmatization\footnote{\url{https://spacy.io/}}. We then use an edit-distance-based string matching library\footnote{\url{https://docs.python.org/3/library/difflib.html}} to search for translation pairs where proverbs are contained in the source sentence. When a source sentence contains an exact match of the given proverb, its matching score will be 1.0, any replacement of characters will reduce the score. We set a threshold as 0.8 during the search.

\subsubsection{Fine-grained Filtering} 
\label{sec:p2_filter} 

Although lemmatization can solve some of the matching errors caused by morphological inflection, the search results will still contain a large number of errors. Therefore we propose two fine-grained methods for further filtering through the semantics of the mined sample.

\vspace{-1.5mm}
\paragraph{LLM-based Proverb Usage Filtering} We use an LLM\footnote{\texttt{meta-llama/Meta-Llama-3.1-70B-Instruct}} to filter those translation pairs that contain the proverb by asking the model: ``Whether the proverb is contained in the sentence". This ensures that the filtering is performed through the semantic meaning of the given text and thus is more accurate. To make sure that the LLM's output is reliable, we set the temperature to 0 and constrain the model's output as ``Yes'' and ``No''. For samples labeled as ``NO'', we remove them from our candidate set.

\vspace{-1.5mm}
\paragraph{Filtering with Quality Estimation} Another filtering process aims to ensure the translation of the collected pairs is good enough, as there are cases where source and target texts are mismatched in the subtitles due to reordering of utterances which have to be excluded. In this step, we use both LLM (LLM-QE) and a dedicated Quality Estimation with Direct Assessment (DA-QE) model to score the collected translation.
\begin{itemize}[noitemsep,topsep=2pt,parsep=2pt,partopsep=2pt]
\item For LLM-QE, we let the model to score the translation from 1-5 and replace the order of source and target, then, score it again to reduce the influence of the order; an average of two values is computed as the label of the candidate pair. 
\item For DA-QE, we use ``Unbabel/wmt23-cometkiwi-da-xxl'' as the dedicated DA scorer to score the translation pair in a range between 0 to 1. Then, we compute the overall score for each pair as $\text{score} = \text{score}_{\text{LLM-QE}} + \text{score}_{\text{DA-QE}} \times 5$. 
\end{itemize}

Filtering is conducted separately across language pairs as the quality of the mined corpus in each direction differs largely. Specifically, we set the maximum required sample size for each direction as \textbf{2000}, and use it to compute the minimum quantile as ($q_{\min}=max(0,1-\frac{2000}{|\mathcal{D}_{s \rightarrow t}|})$, where $|\mathcal{D}_{s \rightarrow t}|$ stands for the number of samples for a specific direction e.g. En$\rightarrow$De) and the corresponding overall score ($max(\text{score}_{q_{\min}}, 4)$, where $\text{score}_{q_{\min}}$ is the corresponding score of the quantile $q_{\min}$, 4 is the minimum score threshold we assigned for all language pairs) as the threshold. Finally, we used this threshold to filter qualified samples. Detailed statistics in each step are presented in \Cref{tab:filter_statistic}.

\vspace{-1.5mm}
\paragraph{Conversation Context Retrieval} While proverbs and sayings are self-contained, they are typically used in conversation. As we aim to study whether the provided context could influence the translation of a sentence containing a proverb, we need to retrieve the prior and proceeding sentences for each filtered translation pair. In this step, we retrieve a maximum of 5 sentences for each direction (prior and proceeding).

\begin{table}[t]
\begin{center}
\scalebox{0.9}{
  \begin{tabular}{l|rr|rr}
  \toprule
  \textbf{} &\multicolumn{2}{c|}{\textbf{Proverbs}} & \multicolumn{2}{c}{\textbf{PiC}} \\
  \textbf{} & \multicolumn{1}{c}{\textbf{\#lit.}} & \multicolumn{1}{c|}{\textbf{\#fig.}} & \multicolumn{1}{c}{\textbf{\#lit.}} & \multicolumn{1}{c}{\textbf{\#fig.}} \\
  \midrule
  en -> bn & 130 & 163  & 29 &13 \\
  bn -> en & 68 & 272& 2 & 6\\
  \midrule
  en -> de & 180 & 214 & 1191 & 349\\
  de -> en & 151& 183  & 614& 515\\
  \midrule
  en -> id & 162 & 183& 886 & 328\\
  id -> en & 71 & 262 & 3 & 10\\
  \midrule
  en -> zh & 134 & 161 & 227 & 45\\
  zh -> en & 191 & 143 & 386 & 441\\
  \bottomrule                 
  \end{tabular}
 }
   \vspace{-2mm}
   \caption{Data statistics of the standalone proverb translation (Proverbs) and proverb in conversation translation (PiC). The number of samples contains literal and figurative proverbs are denoted by \textbf{\#lit} and \textbf{\#fig} respectively.}
   \label{tab:parallel-data}
   \end{center}
   \vspace{-5mm}
\end{table}

\subsubsection{Human Evaluation}
\label{sec:p3_filter}



To validate the collected translation data, the annotators are presented with a proverb in source language, and ask to evaluate the translation in target language given the preceding and following context. Additionally, they are asked whether the proverb is correctly used in the given sentences. The details of annotation protocols and addition analysis can be found in \Cref{appx:data}. Finally, we collect the parallel pairs which contains correct proverb usage and correct translation. The data statistics are shown in \Cref{tab:parallel-data}. Since the sample size of \textsc{En}->\textsc{Bn}, \textsc{Bn}->\textsc{En} and \textsc{Id}->\textsc{En} in our final PiC dataset, we omit these three translation directions in our experiments.
\section{Experimental Setup}
In this paper, we investigate the ability of the current MT system, with a particular focus on LLM-based MT models, in translating proverbs. 
More specifically, we aim to answer the following research questions (\textbf{RQs}):

\begin{itemize}[noitemsep,topsep=2pt,parsep=2pt,partopsep=2pt]
    \item \textbf{RQ1}: To what extent can existing MT methods accurately translate proverbs?
    \item \textbf{RQ2}: What are the roles of conversation contexts and prompts in the proverb translation ability of LLM-based MT?
    \item \textbf{RQ3}: Are current automatic evaluation metrics reliable and effective in measuring the accuracy of proverb translation?
\end{itemize}

\subsection{Models}
We study the proverb translation ability of the current MT methods, including
\begin{itemize}[noitemsep,topsep=2pt,parsep=2pt,partopsep=2pt]
    \item \textbf{State-of-the-art multilingual NMT} We experiment with \nllb which are trained on translation data of 200 languages and available with different sets of parameters: 600M, 1.3B and 3.3B \citep{Costajussa2022Nolanguageleft}.\footnote{Model signatures: \texttt{facebook/nllb-200-distilled-600M}, \texttt{facebook/nllb-200-1.3B}, and \texttt{facebook/nllb-200-3.3B}} 
    \item \textbf{Instruction-following LLMs}  We evaluate instruction-following LLMs with different model parameter size from multiple model families: \mistral (7B)~\citep{jiang2023mistral}, \qwen (7B)~\citep{yang2024qwen2}, \llama (8B, 70B)~\citep{dubey2024llama}, \gemma~\citep{team2024gemma} and \gptfouro.\footnote{Signatures: \texttt{mistralai/Mistral-7B-Instruct-v0.3}, \texttt{Qwen/Qwen2-7B-Instruct}, \texttt{meta-llama/Meta-Llama-3.1-} \texttt{8B-Instruct}, 
 \texttt{meta-llama/Meta-Llama-3.1-70B-Instruct}}
    \item \textbf{LLM-based MT} In addition to NMT and off-the-shelf LLMs models, we also evaluate fine-tuned LLM model for MT tasks. Particularly, we consider \alma which is based on \texttt{Llama2 13B} further fine-tuning with contrastive preference optimization on high quality translation data~\citep{xu2024contrastive}.\footnote{Model signature: \texttt{haoranxu/ALMA-13B-R}.}
\end{itemize}

\begin{table*}[t]
\begin{center}
\scalebox{0.8}{
  \begin{tabular}{l|rr|rr||rr|rr||rr|rr}
  \toprule
   &\multicolumn{4}{|c||}{\textbf{BLEU}} & \multicolumn{4}{c||}{\textbf{CHRF++}} & \multicolumn{4}{c}{\textbf{COMET}} \\
  & \multicolumn{2}{|c|}{lit.} &  \multicolumn{2}{c||}{fig.}
   & \multicolumn{2}{c|}{lit.} &  \multicolumn{2}{c||}{fig.}
   & \multicolumn{2}{c|}{lit.} &  \multicolumn{2}{c}{fig.} \\
   & \multicolumn{1}{|c}{$0$-shot} & \multicolumn{1}{c|}{$1$-shot} & \multicolumn{1}{c}{$0$-shot} & \multicolumn{1}{c||}{$1$-shot} &
   \multicolumn{1}{c}{$0$-shot} & \multicolumn{1}{c|}{$1$-shot} & \multicolumn{1}{c}{$0$-shot} & \multicolumn{1}{c||}{$1$-shot} &
   \multicolumn{1}{c}{$0$-shot} & \multicolumn{1}{c|}{$1$-shot} & \multicolumn{1}{c}{$0$-shot} & \multicolumn{1}{c}{$1$-shot}  \\
  \midrule
  \midrule
  \multicolumn{13}{l}{\quad \quad \quad \quad \quad \quad \quad \quad \textit{From English translation}} \\
\nllbsmall  & 9.42 &  & 8.23 &  & 23.14 &  & 21.49 &  & 67.36 &  & 60.30 & \\
\nllbmedium & 10.66 &  & 8.97 &  & 24.12 &  & 21.88 &  & 67.94 &  & 60.37 & \\
\nllbbig & 11.23 &  & 9.17 &  & 24.73 &  & 22.43 &  & 67.99 &  & 60.60 & \\
\midrule
\alma & 6.55 &  & 5.68 &  & 17.97 &  & 16.93 &  & 62.55 &  & 55.56 & \\
\qwen & 9.57 & \underline{10.37} & \underline{8.17} & 7.94 & 22.42 & \underline{22.93} & \underline{20.71} & 20.70 & 67.13 & \underline{67.86} & 60.55 & \underline{60.86} \\
\mistral & 6.44 & \underline{6.92} & \underline{5.24} & 5.06 & \underline{18.91} & 18.74 & \underline{16.79} & 16.57 & 63.79 & \underline{64.46} & 56.07 & \underline{56.86} \\
\gemma & 13.27 & \underline{14.22} & 10.23 & \underline{10.40} & 24.89 & \underline{25.54} & 22.52 & \underline{22.69} & \underline{71.13} & 70.88 & 63.32 & \underline{63.38} \\
\llamasmall & 10.51 & \underline{11.08} & 8.22 & \underline{8.87} & 23.40 & \underline{23.68} & 20.67 & \underline{21.33} & 69.44 & \underline{70.17} & 62.28 & \underline{62.76} \\
\llamalarge & \textbf{15.16} & \underline{\textbf{16.83}} & \textbf{13.53} & \underline{\textbf{13.97}} & \textbf{27.59} & \underline{\textbf{28.68}} & \textbf{25.59} & \underline{\textbf{26.12}} & \textbf{72.49} & \underline{\textbf{73.09}} & \textbf{65.24} & \underline{\textbf{65.55}} \\
\gptfouro & \underline{13.61} & 12.82  & \underline{11.20} & 10.64 & \underline{26.97} & 26.62 & \underline{24.61} & 24.04 & 71.62 & \underline{71.65} & \underline{63.45} & 63.37 \\
\midrule
\midrule
\multicolumn{13}{l}{\quad \quad \quad \quad \quad \quad \quad \quad \textit{To English translation}} \\
\nllbsmall  & 11.42 &  & 11.42 &  & 28.63 &  & 28.35 &  & 60.79 &  & 57.91 & \\
\nllbmedium & 14.30 &  & 14.12 &  & 30.83 &  & 30.70 &  & 62.37 &  & 59.38 & \\
\nllbbig & 15.84 &  & 14.41 &  & 31.41 &  & 31.35 &  & 62.71 &  & 59.69 & \\
\midrule
\alma & 13.98 &  & 17.42 &  & 29.45 &  & 32.48 &  & 61.33 &  & 61.19 & \\
\qwen & \underline{15.66} & 15.07 & 16.47 & \underline{16.66} & 32.25 & \underline{32.26} & \underline{33.79} & 33.50 & \underline{66.46} & 66.24 & \underline{63.45} & 63.23 \\
\mistral & 13.03 & \underline{13.58} & \underline{14.83} & 14.46 & 27.98 & \underline{28.91} & \underline{30.47} & 30.34 & 61.75 & \underline{62.75} & 60.39 & \underline{60.89} \\
\gemma & \underline{17.78} & 17.43 & \underline{18.93} & 18.41 & \underline{34.97} & 34.30 & \underline{35.98} & 35.33 & \underline{67.78} & 67.50 & \underline{64.35} & 64.21 \\
\llamasmall & 15.66 & \underline{16.72} & 16.20 & \underline{16.77} & 30.84 & \underline{31.67} & 32.03 & \underline{32.69} & 64.89 & \underline{65.64} & 62.37 & \underline{63.24} \\
\llamalarge & \underline{\textbf{20.85}} & \textbf{20.10} & \underline{\textbf{22.35}} & \textbf{21.55} & \underline{36.73} & \textbf{36.57} & \underline{\textbf{38.22}} & \textbf{37.26} & \underline{68.77} & \textbf{68.74} & \underline{\textbf{65.29}} & \textbf{65.00}\\
\gptfouro & \underline{18.84} & 17.14 & \underline{18.47} & 18.22 & \textbf{37.15} & 36.83 & 38.05 & \underline{38.10} & \textbf{68.88} & \underline{69.02} & 65.20 & \underline{65.42} \\
  \bottomrule                 
  \end{tabular}
 }
   \caption{Results on standalone proverb translation. \textbf{Bold} highlights the best score in each column. The better score among one-shot and zero-shot are \underline{underlined}.
   }
   \label{tab:proverb-result}
   \end{center}
   \vspace{-5mm}
\end{table*}

\subsection{Prompts}
We design 5 types of prompt templates to evaluate the performance of LLM under different conditions (see \Cref{fig:subtitle_trans_prompt_template} in the Appendix). All 5 prompt templates are used in the evaluation on the PiC test set, but only the zero- and one-shot prompts are used in proverb standalone translation.
\begin{itemize}[noitemsep,topsep=2pt,parsep=2pt,partopsep=2pt]
    \item \textbf{Zero-Shot} We only provide a simple system message to set a role (a professional translator) for the LLM and instruct it to translate the given source sentence without returning any irrelevant content.
    \item \textbf{One-Shot} As smaller LLMs may have limited instruction-following capability, 
    we add an example of translating the sentence 
    ``Good morning" in the first round of dialogue.
    This allows LLM to have access to the dialogue history.
    \footnote{We use the simple sentence for the one-shot case to prevent introducing any bias to the actual translation.}
    \item \textbf{Proverb Explanation} 
    In the system message, 
    we first signal the LLM that the proverb may contained in the given source text, and the explanation of the proverb,
    followed by the source sentence.
    \item \textbf{Contextualization through Dialogue} 
    To study the role of the conversation context in
    translation performance, we consider previous subtitle sentences as contexts and place them in the dialogue history, up to 5 rounds with source and target sentences acting as user input and model responses (essentially becoming a maximum of 5-shot form).
    \item \textbf{Contextualization through Concatenation} Although using a dialogue format to provide previous sentences as contexts to the model is an intuitive approach, it may introduce noise when source and target sentence pairs in the context containing reordering. To this end, we design another approach for contextualization by concatenating all source and target sentence pairs in the context into one user input and model response, making it into a one-shot form. Through this, reordered contexts can be placed in the same round and naturally recovered to the correct alignment.
\end{itemize}

\vspace{-2mm}
\subsection{Evaluation}
\vspace{-1.5mm}
\paragraph{Evaluation Metrics} We evaluate the translation quality with lexical-overlap metrics including BLEU~\citep{papineni-etal-2002-bleu} and CHRF++~\citep{popovic-2017-chrf} using SacreBLEU~\citep{post-2018-call},
and neural evaluation metric such as COMET~\citep{rei-etal-2020-comet}.\footnote{COMET signature: \texttt{Unbabel/wmt22-comet-da}.}
\vspace{-1.5mm}
\paragraph{Inference}  We use sampling with the default decoding parameters for \gptfouro, and beam search with a beam size of 5 for \nllb and open-source LLMs.


\section{Main Results}

\subsection{Standalone Proverb Translation}
\vspace{-1.5mm}
\paragraph{NMT vs LLMs} \Cref{tab:proverb-result} presents the performance of various models on both literal and figurative proverb translation with zero-shot and one-shot prompting. Notably, despite being a LLM specialized for MT, \alma consistently underperforms compared to other LLMs across the board and even falls behind the smallest \nllb model in from-English translation direction. We speculate that it can be attributed to the absence of Bengali and Indonesian languages in its fine-tuning data. Similarly, \mistral also struggles on this task due to its limited language support and relatively smaller size. In contrast, other LLMs outperform \nllb models, with \llama 70B emerging as the strongest model.

\vspace{-1.5mm}
\paragraph{Zero-shot vs One-shot Prompting} One-shot prompting generally outperforms zero-shot prompting, especially in  from-English translation direction. Interestingly, our strongest model \llama 70B achieves slightly higher score in zero-shot prompting in to-English translation tasks. 

\vspace{-1.5mm}
\paragraph{Literal vs Figurative Proverbs} Overall, the performance on figurative proverbs  is consistently lower than on literal proverbs across all metrics in from-English translation directions.  It is expected as the figurative proverbs have an underlying meaning different to their literal wording which makes them more challenging to translate. However, in the to-English translation direction, we notice an unexpected trend across different metrics.
While \nllb models generally score higher on literal proverbs than figurative one for all metrics, the opposite trend is observed with LLMs. Specifically, LLMs achieve higher BLEU and CHRF++ scores when translating figurative proverbs, but they tend to score better COMET scores on literal proverbs. 

\begin{figure}
\vspace{-1em}
\begin{subfigure}
        \centering
        \includegraphics[scale=0.4]{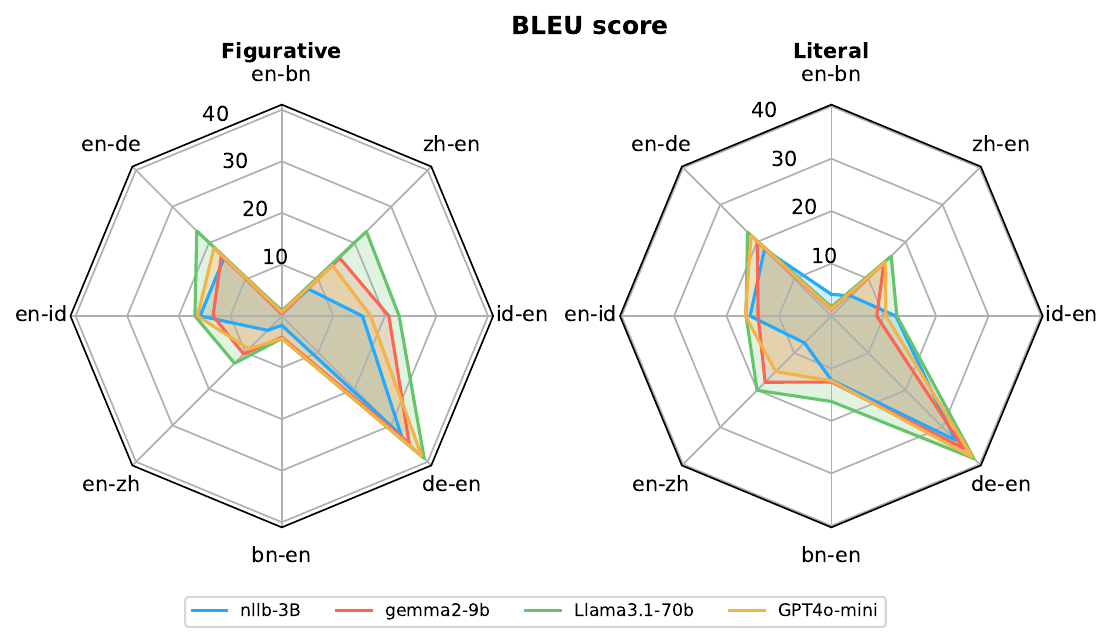}
    \end{subfigure}%
    \begin{subfigure}
        \centering
        \includegraphics[scale=0.4]{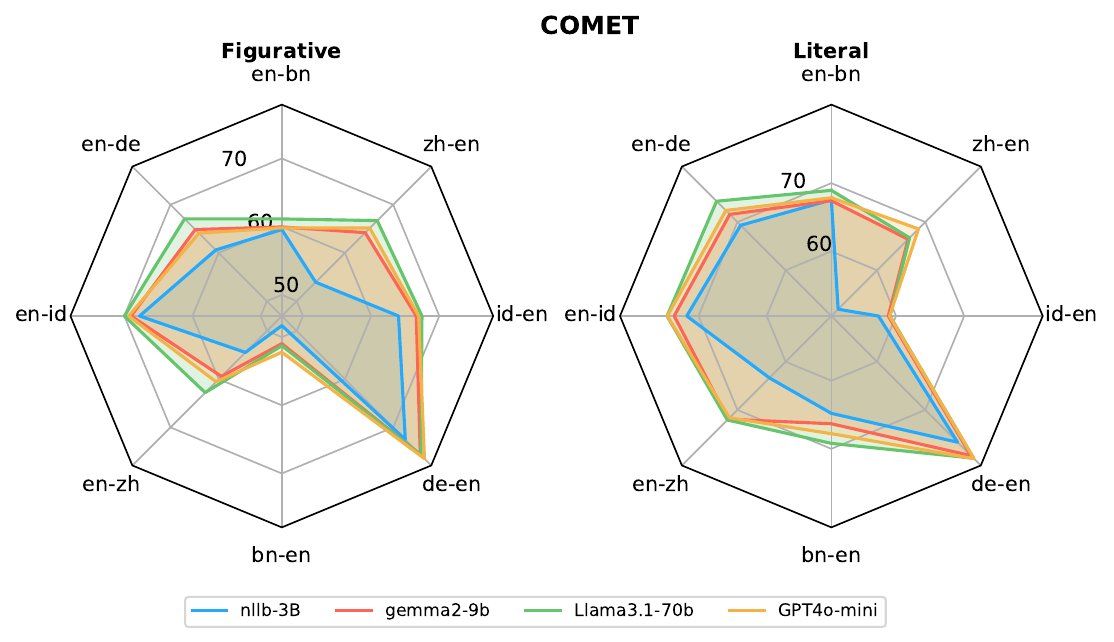}
    \end{subfigure}
\caption{Result on proverb translation of each translation direction.}
     \label{fig:proverb-radar}
    \vspace{-1em}
\end{figure}

\Cref{fig:proverb-radar} breaks down the performance of 4 models on literal and figurative proverb translation in each translation direction. All models perform reasonably well in \textsc{De-En} pairs because both languages are high-resource and belong to similar cultural regions. On the other hand, \textsc{Bn-En} are the most challenging tasks. Overall, all models perform better on literal translation in all translation directions, except the \textsc{Id->En} direction. This leads to the average performance on to-English figurative proverb translation is higher than literal ones.




\begin{table*}[t]
\begin{center}
\scalebox{0.8}{
  \begin{tabular}{l|rrrrr|rrrrr}
  \toprule
   &\multicolumn{5}{|c}{\textbf{Literal}} & \multicolumn{5}{|c}{\textbf{Figurative}} \\
   & \multicolumn{1}{|c}{$0$-shot} & \multicolumn{1}{c}{$1$-shot} & \multicolumn{1}{c}{\textsc{Expl.}} & \multicolumn{1}{c}{\textsc{Dialog}} &
   \multicolumn{1}{c}{\textsc{Concat}} & \multicolumn{1}{|c}{$0$-shot} & \multicolumn{1}{c}{$1$-shot} & \multicolumn{1}{c}{\textsc{Expl.}} & \multicolumn{1}{c}{\textsc{Dialog}} &
   \multicolumn{1}{c}{\textsc{Concat}}   \\
  \midrule
  \midrule
  \multicolumn{11}{l}{\quad \quad \quad \quad \quad \quad \quad \quad \textit{From English translation, incl. \textsc{En-De}, \textsc{En-Id}, \textsc{En-Zh}}} \\
\nllbsmall  & 84.93 &  &  &  &  & 78.87 &  &  &  &  \\
\nllbmedium & 86.04 &  &  &  &  & 80.18 &  &  &  & \\
\nllbbig & 85.34 &  &  &  &  & 80.32 &  &  &  & \\
 \midrule
\alma & 84.94 &  &  &  &  & 80.41 &  &  &  & \\
\qwen & 84.50 & 84.95 & 84.31 & \underline{85.74} & \underline{85.74} & 80.63 & 81.40 & 81.24 & \underline{83.08} & 82.71 \\
\mistral & 81.63 & 81.54 & 81.48 & \underline{83.38} & 82.88 & 76.37 & 76.10 & 76.66 & \underline{77.94} & 77.34 \\
\gemma & 85.51 & 85.31 & 85.12 & \underline{86.19} & 86.05 & 82.05 & 82.20 & 82.54 & \underline{82.82} & 82.62 \\
\llamasmall & 83.72 & 84.69 & 83.30 & \underline{85.75} & 85.59 & 79.16 & 78.82 & 78.37 & \underline{80.96} & 80.57 \\
\llamalarge & 86.29 & 86.42 & 86.37 & \underline{87.30} & 86.71 & 84.26 & 83.92 & \textbf{84.81} & \underline{85.14} & 84.44 \\
\gptfouro & \textbf{87.22} & \textbf{87.36} &\textbf{86.91} & \textbf{\underline{88.06}} & \textbf{87.95} & \textbf{84.60} & \textbf{84.70} & 84.75 & \textbf{\underline{85.59}} & \textbf{84.61} \\
   \midrule
  \midrule
  \multicolumn{11}{l}{\quad \quad \quad \quad \quad \quad \quad \quad \textit{To English translation, incl. \textsc{De-En}, \textsc{Zh-En}}} \\
  \nllbsmall  & 60.07 &  &  &  &  & 61.61 &  &  &  & \\
\nllbmedium & 63.16 &  &  &  &  & 63.17 &  &  &  &  \\
\nllbbig & 62.99 &  &  &  &  & 63.64 &  &  &  & \\
 \midrule
\alma & 64.83 &  &  &  &  & 65.18 &  &  &  & \\
\qwen & 64.32 & 64.04 & 64.78 & \underline{65.86} & 65.45 & 66.41 & 64.98 & 66.26 & \underline{68.64} & 67.74 \\
\mistral & 60.54 & 62.86 & 61.33 & \underline{64.83} & 64.15 & 63.14 & 62.46 & 62.88 & \underline{65.90} & 65.10 \\
\gemma & 64.87 & 65.93 & 66.05 & \underline{67.83} & 67.96 & 66.22 & 66.98 & 66.81 & \underline{68.98} & 68.37 \\
\llamasmall & 62.83 & 65.17 & 62.67 & \underline{67.22} & 66.53 & 63.98 & 65.55 & 65.15 & \underline{66.44} & 66.21 \\
\llamalarge & 65.76 & 66.28 & \textbf{66.93} & \textbf{\underline{68.83}} & 68.01 & 65.59 & 67.86 & \textbf{67.25} & \underline{68.48} & 68.44 \\
\gptfouro & \textbf{66.30} & \textbf{66.32} & 66.30 & \underline{68.30} & \textbf{68.17} & \textbf{66.65} & \textbf{68.03} & 66.79 & \textbf{\underline{68.87}} & \textbf{68.46} \\
  \bottomrule                 
  \end{tabular}
 }
   \caption{Subtitle Translation (COMET score) beam search. \textbf{Bold} highlights the best score in each column. The better score among different prompting methods is \underline{underlined}.
   }
   \label{tab:comet-bleu}
   \end{center}
   \vspace{-5mm}
\end{table*}

\subsection{Proverb in Conversation Translation}
\vspace{-1.5mm}
\paragraph{NMT vs LLMs}\Cref{tab:comet-bleu} reports COMET scores of various models on the translation of proverbs in conversational contexts. Detailed results in BLEU and CHRF++ can be found in the Appendix.
Consistent with earlier findings in proverb translation, \mistral 7B lags behind other models, while \gptfouro stands out as the strongest LLM model, following by \llama 70B.
 In this particular task, interestingly, \nllb models show highly competitive results to the LLMs, especially in the literal proverb subset. However, \nllb models fall short in translating figurative proverbs.

\vspace{-1.5mm}
 \paragraph{Roles of Context and Prompting} Among the different prompting strategies, one-shot prompting consistently improves upon the performance of zero-shot prompting. However, we do not observe any notable improvement when providing explanations of proverbs, particularly in the case of literal proverbs. This may be because LLMs have already exposed to the meanings of  proverbs during their pre-training, allowing them to capture these meanings without additional explanation.
 On the other hand, incorporating conversational context significantly enhances translation performance. 
 This is likely due to the extra context clues in the conversations, which help the models generate more accurate translations.
Furthermore, framing the context as a dialogue proves to be more effective than simple concatenation, as it aligns more naturally with the conversational nature of LLMs.


\begin{figure*}[t]
    \centering
    \begin{subfigure}
    \centering
    \footnotesize
    \includegraphics[width=1.0\textwidth]{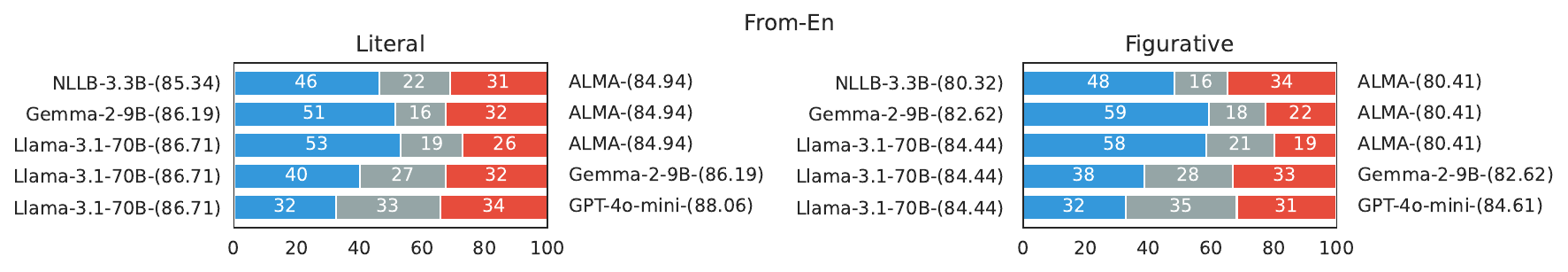}
    \label{fig:winrate-from-en}
    \end{subfigure}%
    \vspace{-1.5em}
    \begin{subfigure}
    \centering
    \footnotesize
    
    \includegraphics[width=1.0\textwidth]{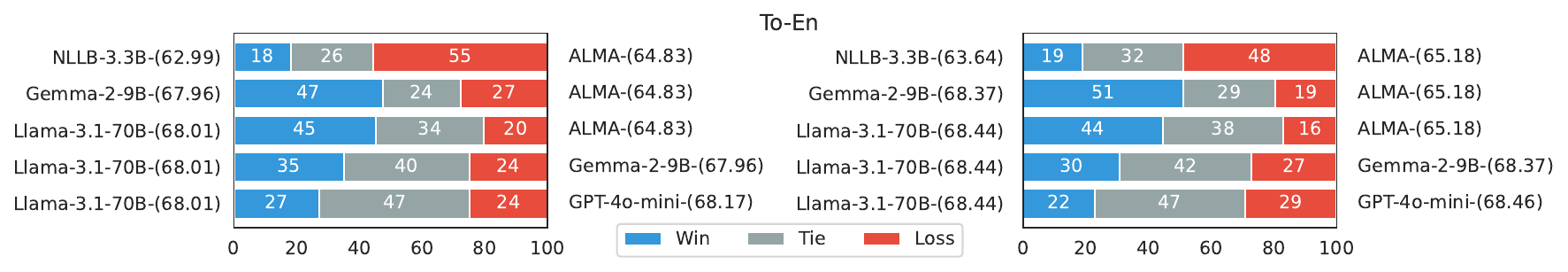}
    \label{fig:winrate-to-en}
    \end{subfigure}%
    \vspace{-2em}
    \caption{We present the win rate of 5 pairs of models evaluated by the \gptfouro. Results are separated with From/To-En and Figurative/Literal, with corresponding COMET scores indicated near the model name.}
    \label{fig:winrate}
\vspace{-3mm}
\end{figure*}

\section{Analysis}
\subsection{Limitations of Evaluation Metrics on Proverb Translation}
\begin{table*}[t!]
\begin{center}
\scalebox{0.8}{
\footnotesize
\renewcommand{\arraystretch}{1.3}
\begin{tabular}{|l|c|c|c|}
\hline
\textbf{Example} & \textbf{BLEU} & \textbf{CHRF} & \textbf{COMET} \\
\hline
\hline
\multicolumn{4}{|l|}{\textbf{Reference:} ``Distance determines the stamina of a horse.''} \\
\hline
\textbf{Hypothesis 1:} ``Distance reveals the strength of a horse'' & 26.27 & 46.19 & 82.85 \\
\textbf{Hypothesis 2:} ``\textcolor{red}{A long journey} reveals the strength of a horse'' & 19.06 & 35.42 & 71.58 \\
\hline
\hline
\multicolumn{4}{|l|}{\textbf{Reference:} ``The face of a tiger, the heart of a mouse.''} \\
\hline
\textbf{Hypothesis 1:} ``The face of a tiger, the heart of a mouse'' & 89.32 & 95.05 & 93.23 \\
\textbf{Hypothesis 2:} ``The look of a tiger, the heart of a \textcolor{red}{rat}.'' & 71.03 & 80.81 & 81.99 \\
\hline
\hline
\multicolumn{4}{|l|}{\textbf{Reference:} ``Spare the rod and spoil the child.''} \\
\hline
\textbf{Hypothesis 1:} ``Spare the rod and spoil the child.'' & 100.00 & 100.00 & 95.22 \\
\textbf{Hypothesis 2:} ``\textcolor{red}{Discipline brings forth filial} children.'' & 0.0 & 13.93 & 60.10 \\
\hline
\hline
\end{tabular}
}
\caption{Evaluation of Hypotheses with BLEU, CHRF, and COMET Scores. \textcolor{red}{Red} highlights the words that make the metrics score the hypotheses lower.}
\label{tab:metric-analysis}
  \end{center}
\vspace{-6mm}
\end{table*}
As proverbs are often highly culture-specific, standard evaluation metrics like BLEU, CHRF, or COMET may fall short in certain cases. In this section, we illustrate the weaknesses of these metrics regarding proverb translation. With all of the hypotheses from different models and prompt templates, we compute the cosine similarity between the sentence embedding\footnote{We use \texttt{sentence-transformers/all-mpnet-base-v2} model to obtain the embeddings.} of the hypothesis and its reference.
We focus on cases where the cosine similarity difference of two hypotheses against the same reference is low ($< 0.05$), while the difference is above a threshold of $10.0$, $5.0$ and $10.0$ for COMET, BLEU and CHRF++, respectively.
From the total of 5M hypothesis pairs, we find that 22,704 pairs satisfy these thresholds. Notably, specialized NMT models (\nllb and \alma) have the fewest appearances in these cases, accounting for 1,962 pairs. This may indicate that LLMs may 
produce more creative translations that the evaluation metrics may have neglected.
\vspace{-1.5mm}
\paragraph{Qualitative Analysis}
We manually check the detected cases to identify problems of evaluation metrics. \Cref{tab:metric-analysis} shows some cases where the metrics are unreliable for evaluating proverb translation. 
In the first and second examples, the hypothesis 2 scores lower on all metrics due to the usage of different phrases to the reference, even though it conveys the same idiomatic meaning ("a long journey" instead of "distance", and "rat" instead of "mouse"). This indicates that the metrics are overly sensitive to surface-level lexical differences.
The third example, "Spare the rod and spoil the child," showcases the metrics' inability to handle significant paraphrasing. 
Despite conveying the same core meaning, the metrics fail to recognize the equivalence due to their reliance on word overlap rather than deeper semantic understanding. These examples demonstrate that standard metrics can be inadequate when evaluating translations involving idioms and non-literal expressions, as they tend to penalize valid translations that do not strictly adhere to the reference's lexical choices. Even COMET, while more robust in capturing meaning, struggles with cases of metaphorical language and significant rephrasing. 

\subsection{LLM-as-a-Judge Evaluation}

Realizing the ineffectiveness of traditional evaluation metrics, we further use the \textbf{LLM-as-a-judge} method for evaluation. Here, we primarily evaluate the PiC translation results and follow the setup of \Cref{tab:comet-bleu}.
We selected five representative model pairs and compared their translation results using \gptfouro, calculating the win rates.
For \nllb and \alma, we use zero-shot results, while for other LLMs, we use results generated with dialogue prompts.
The prompt used for evaluation is shown in \Cref{fig:llm_eval_prompt_template} in the Appendix. We prompted the model to compare the translations comprehensively based on three aspects: translation accuracy, fluency, and cultural appropriateness. Additionally, we provided the complete contextual information before and after the sample, along with the corresponding reference to assist the model in its judgment. To avoid the bias caused by the positioning of hypothesis A and B, we randomly assigned the results of the two models as A and B, thus eliminating the influence of position. Finally, \gptfouro will return one of three results: A, B, or tie.

In \Cref{fig:winrate}, we present the win rates along with the COMET scores of the two groups of models. It can be seen that the win rates and COMET scores generally follow a consistent trend (models with higher COMET scores usually have higher win rates). However, the gap in COMET scores does not strictly correlate linearly with the win rate. For instance, a larger difference in COMET scores does not necessarily mean a higher win rate (e.g. \llamalarge vs \alma, \llamalarge vs \gemma). This suggest that  evaluation with LLM-as-judge cannot fully solve the limitations of traditional evaluation methods.

\subsection{Data Contamination Analysis}
\label{sec:contam}

\begin{table}[]
\centering
\resizebox{0.8\columnwidth}{!}{%
\begin{tabular}{@{}l|ccccc@{}}
\toprule
 & \multicolumn{5}{c}{\% samples with $\gamma > 0.9$} \\ \midrule
Models & \textit{En} & Bn & De & Id & Zh \\ \midrule
\llamasmall & 4.5 & 0.0 & 3.8 & 1.1 & 3.8 \\
\gemma & 1.3 & 0.0 & 1.9 & 0.4 & 1.7 \\
\qwen & 4.7 & 0.0 & 1.5 & 0.1 & 1.2 \\
\mistral & 1.0 & 0.0 & 0.7 & 0.3 & 0.8 \\ \bottomrule
\end{tabular}%
}
\caption{In this table, we present the percentage of samples with $\gamma>0.9$ as the measurement of contamination. 
}
\label{tab:contamination_rate}
\vspace{-3mm}
\end{table}
Following \citep{Liu2024AreMultilingualLLMs}, we measure the memorization rate of the Proverb-in-Conversation dataset by assessing the model ability to complete an utterance based on partial context.  This is quantified using the longest common subsequence (LCS) rate between the model's prediction and the actual utterance, given the preceding context. However, since proverbs are often well-known phrases, models might accurately predict them even without relying on the provided context. To account for this, we introduce the contamination rate $\gamma$ as the difference between the LCS of the model’s prediction and the reference with and without context, normalized by the utterance length.  We provide more details of this metric in \Cref{appx:data-contamination}.

A higher value of $\gamma$ indicates a greater likelihood that the model has been affected by data contamination for a given sample. We present the percentage of samples with $\gamma > 0.9$ in Table \ref{tab:contamination_rate} across each language. A relatively higher correlation between the contamination rate and the resource-level of the language can be found, but the correlation to the translation performance is not significant. This suggests that the contamination issue is not biasing our evaluation.


\section{Conclusion}
\vspace{-1.5mm}
We curate a proverb and its usage in conversation to investigate the ability of LLMs on proverb translation. Our experiments reveal that LLMs generally outperform NMT model on this task, showcasing the advantage of LLMs in translating figurative expressions, especially between high-resource languages and languages from similar cultural region. Additionally, our analysis also reveals that current automatic evaluation metrics are unreliable in measuring translation with figurative languages. 

\section*{Limitation}
Although we have constructed the Proverb in Conversation dataset and conducted systematic evaluations of different models, we acknowledge the following two limitations of this study:
\begin{itemize}
    \item The scale of our dataset is currently relatively small. This is partly due to the limited number of proverbs in each language, which constrains the size of the samples that can be collected. Additionally, our strict filtering and human annotation process further excluded low-quality samples, which may have led to excessive discarding and thus limited the dataset size. In future work, we will consider expanding data sources and ensure that the scale of the collected data meets the needs for more comprehensive evaluations.
    \item Another limitation lies in the relatively limited variety of models and prompts tested. Although most of them are commonly used models, they still cannot fully represent the capabilities of other models, especially those with more than 70B or fewer than 7B parameters. Therefore, in our future work, we will further expand the selection of models and the variety of prompts to enhance the comprehensiveness of the evaluation.
\end{itemize}


\bibliography{latex/c2mt}

\newpage
\appendix
\appendix
\label{sec:appendix}
\section{Related works}
\paragraph{Figurative Expression in MT} 
Multi-word expressions (MWEs), including idioms, phrasal verbs, and multi-word named entities, present a unique challenge in natural language processing due to their non-compositional nature, where the meaning of the whole expression cannot be easily inferred from the meanings of its individual words~\citep{Constant2017CoLI}. 
Previous research has largely focused on understanding and paraphrasing MWEs, particularly in English~\citep{liu-hwa-2016-phrasal,wada-etal-2023-unsupervised}. In NMT literature, much focuses has been on idiomatic and slang translation, primarily targeting European languages~\citep{fadaee-etal-2018-examining,Sun2022SemanticallyInformedSlang,baziotis-etal-2023-automatic}. 
However, a key obstacle to further progress in this area is the absence of standardized evaluation benchmarks and metrics. Our work addresses this gap by constructing a dataset on the translation of proverbs for four language pairs, each representing distinct geographical and cultural regions.

\paragraph{LLM-based MT} Several studies have explored the application of LLMs for translation tasks, highlighting their impressive performance across multiple high-resource language pairs~\citep{xu2024a,xu2024contrastive,wu2024adapting}. One notable advantage of LLMs over traditional neural machine translation (NMT) systems is their ability to generate more controlled and nuanced translations, particularly when dealing with idiomatic expressions that require less literal interpretation~\citep{manakhimova-etal-2023-linguistically,stap-etal-2024-fine}. In this work, we focus on evaluating the capabilities of LLMs in proverb translation, a challenging task due to the cultural and figurative nature of proverbs.
\section{Data Annotation}
\label{appx:data}
We use Label Studio\footnote{\url{https://labelstud.io/}} as our annotation platform. For Bengali, Indonesian and Mandarin Chinese, we recruit four annotators per language. For German-English, two annotators are recruited.

\begin{figure*}
    \centering
    \includegraphics[scale=0.4]{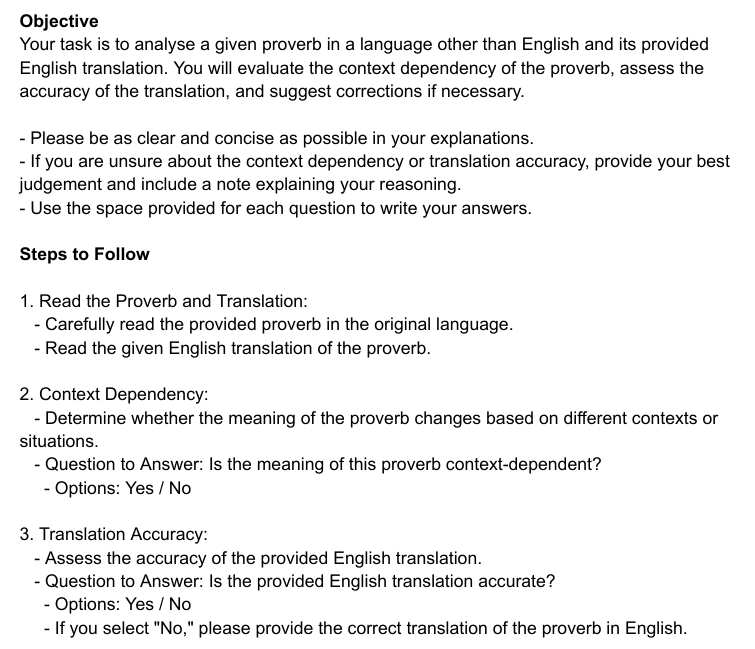}
    \caption{Proverb Translation Annotation Instruction}
    \label{fig:proiverb-inst}
\end{figure*}

\begin{figure*}
    \centering
    \includegraphics[scale=0.4]{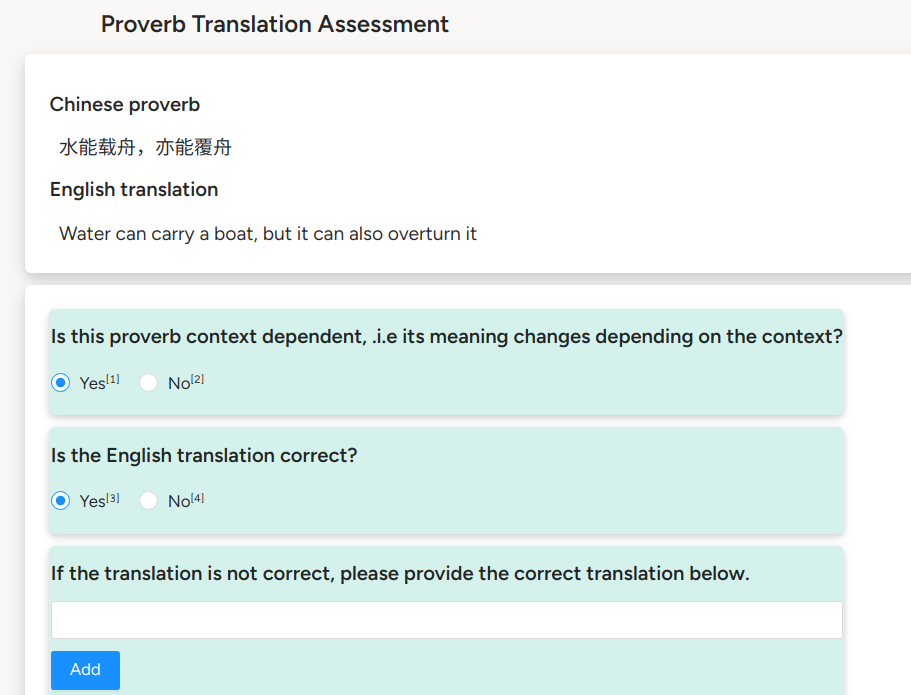}
    \caption{Proverb Translation Annotation Interface}
    \label{fig:proiverb-ui}
\end{figure*}

\Cref{fig:proiverb-inst} shows the instruction to annotate proverb translation dataset. \Cref{fig:proiverb-ui} shows the Label Studio interface for the task.

\begin{figure*}
    \vspace{-1em}
    \centering
    \includegraphics[scale=0.4]{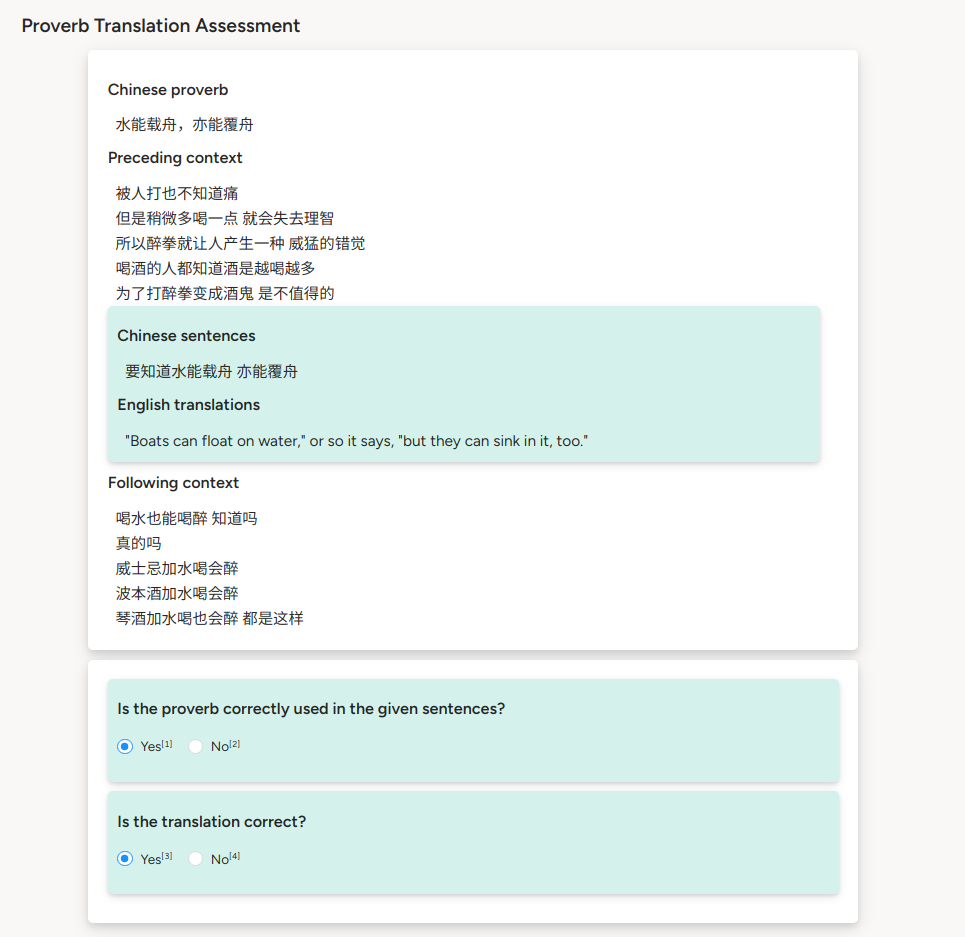}
    \vspace{-1em}
    \caption{OpenSubtitle Translation Annotation Interface}
    \label{fig:subtitle-ui}
\end{figure*} 
\section{Additional Experiment Details}
\begin{table*}[t]
\begin{center}
\scalebox{0.8}{
  \begin{tabular}{l|rrrrr|rrrrr}
  \toprule
   &\multicolumn{5}{|c}{\textbf{Literal}} & \multicolumn{5}{|c}{\textbf{Figurative}} \\
   & \multicolumn{1}{|c}{$0$-shot} & \multicolumn{1}{c}{$1$-shot} & \multicolumn{1}{c}{\textsc{Expl.}} & \multicolumn{1}{c}{\textsc{Dialog}} &
   \multicolumn{1}{c}{\textsc{Concat}} & \multicolumn{1}{|c}{$0$-shot} & \multicolumn{1}{c}{$1$-shot} & \multicolumn{1}{c}{\textsc{Expl.}} & \multicolumn{1}{c}{\textsc{Dialog}} &
   \multicolumn{1}{c}{\textsc{Concat}}   \\
  \midrule
  \midrule
  \multicolumn{11}{l}{\quad \quad \quad \quad \quad \quad \quad \quad \textit{From English translation, incl. \textsc{En-De}, \textsc{En-Id}, \textsc{En-Zh}}} \\
    \nllbsmall  & 27.33 &  &  &  &  & 27.37 &  &  &  & \\
\nllbmedium & 29.85 &  &  &  &  & 29.38 &  &  &  & \\
\nllbbig & \textbf{30.70} &  &  &  &  & 31.01 &  &  &  & \\
\midrule
\alma & 20.14 &  &  &  &  & 23.05 &  &  &  & \\
\qwen & 23.63 & 24.87 & 22.11 & \underline{27.31} & 26.89 & 26.12 & 25.88 & 26.00 & \underline{30.72} & 29.08 \\
\mistral & 17.87 & 18.66 & 16.12 & \underline{22.33} & 22.09 & 19.80 & 19.74 & 18.70 & \underline{22.23} & 21.99 \\
\gemma & 27.93 & 28.24 & 26.85 & \underline{30.94} & 30.88 & 32.01 & 31.08 & 31.37 & \underline{33.81} & 33.17 \\
\llamasmall & 23.85 & 26.62 & 23.26 & 28.99 & \underline{29.13} & 25.64 & 26.09 & 24.98 & \underline{29.45} & 29.05 \\
\llamalarge & 28.59 & 29.19 & \textbf{28.68} & \underline{\textbf{33.30}} & 32.40 & 35.61 & \textbf{35.72} & \textbf{35.07} & \underline{\textbf{37.97}} & \textbf{36.54} \\
\gptfouro & 29.92 & \textbf{30.37} & 27.86 & 32.34 & \underline{\textbf{32.57}} & \textbf{34.27} & 34.51 & 33.91 & \underline{36.47} & 35.03 \\
   \midrule
  \midrule
  \multicolumn{11}{l}{\quad \quad \quad \quad \quad \quad \quad \quad \textit{To English translation, incl. \textsc{De-En}, \textsc{Zh-En}}} \\
\nllbsmall  & 13.96 &  &  &  &  & 18.96 &  &  &  &  \\
\nllbmedium & 18.28 &  &  &  &  & 23.05 &  &  &  &  \\
\nllbbig & \textbf{19.31} &  &  &  &  & \textbf{24.16} &  &  &  &  \\
\midrule
\alma & 16.99 &  &  &  &  & 22.42 &  &  &  &  \\
\qwen & 14.77 & 15.56 & 13.51 & \underline{20.05} & 18.06 & 22.27 & 22.44 & 19.85 & \underline{26.03} & 24.48 \\
\mistral & 12.23 & 13.40 & 11.55 & \underline{19.76} & 17.89 & 18.56 & 19.18 & 17.52 & \underline{23.45} & 21.64 \\
\gemma & 18.17 & 19.23 & 17.19 & \underline{24.51} & 23.66 & 22.97 & 23.34 & 22.40 & \underline{27.29} & 26.66 \\
\llamasmall & 16.52 & 18.42 & 16.04 & \underline{23.65} & 22.48 & 20.31 & 20.89 & 20.91 & \underline{24.89} & 24.54 \\
\llamalarge & 19.28 & \textbf{20.25} & \textbf{19.56} & \underline{\textbf{25.96}} & \textbf{24.07} & 23.53 & \textbf{24.13} & \textbf{23.89} & \underline{\textbf{28.45}} & \textbf{28.08} \\
\gptfouro & 18.07 & 17.78 & 17.25 & \underline{22.48} & 21.88 & 22.72 & 22.83 & 21.68 & \underline{26.08} & 24.89 \\
  
  \bottomrule                 
  \end{tabular}
 }
   \caption{BLEU scores on Proverb in Conversation Translation. \textbf{Bold} highlights the best score in each column. The better score among different prompts are \underline{underlined}.
   }
   \label{tab:subtitle-bleu}
   \end{center}
\end{table*}

\begin{table*}[t]
\begin{center}
\scalebox{0.8}{
  \begin{tabular}{l|rrrrr|rrrrr}
  \toprule
   &\multicolumn{5}{|c}{\textbf{Literal}} & \multicolumn{5}{|c}{\textbf{Figurative}} \\
   & \multicolumn{1}{|c}{$0$-shot} & \multicolumn{1}{c}{$1$-shot} & \multicolumn{1}{c}{\textsc{Expl.}} & \multicolumn{1}{c}{\textsc{Dialog}} &
   \multicolumn{1}{c}{\textsc{Concat}} & \multicolumn{1}{|c}{$0$-shot} & \multicolumn{1}{c}{$1$-shot} & \multicolumn{1}{c}{\textsc{Expl.}} & \multicolumn{1}{c}{\textsc{Dialog}} &
   \multicolumn{1}{c}{\textsc{Concat}}   \\
  \midrule
  \midrule
  \multicolumn{11}{l}{\quad \quad \quad \quad \quad \quad \quad \quad \textit{From English translation, incl. \textsc{En-De}, \textsc{En-Id}, \textsc{En-Zh}}} \\
\nllbsmall  & 40.39 &  &  &  &  & 41.81 &  &  &  & \\
\nllbmedium & 42.45 &  &  &  &  & 43.31 &  &  &  & \\
\nllbbig & \textbf{43.66} &  &  &  &  & 44.17 &  &  &  & \\
\alma & 34.29 &  &  &  &  & 37.36 &  &  &  & \\
\qwen & 37.51 & 38.27 & 36.39 & \underline{40.93} & 39.43 & 40.40 & 40.18 & 40.05 & \underline{46.46} & 41.46 \\
\mistral & 31.65 & 31.71 & 31.74 & \underline{36.79} & 35.52 & 34.60 & 34.34 & 33.94 & \underline{40.53} & 40.16 \\
\gemma & 39.74 & 39.47 & 38.78 & \underline{43.54} & 41.76 & 44.59 &\underline{44.27} & 42.58 & 44.97 & 44.56 \\
\llamasmall & 36.24 & 38.35 & 35.62 & 41.26 & \underline{41.38} & 38.76 & 39.91 & 38.64 & \underline{42.02} & 41.35 \\
\llamalarge & 41.38 & 41.41 & 41.23 & \underline{45.15} & 43.74 & \textbf{46.77} & \textbf{46.79} & \textbf{46.61} & \textbf{\underline{47.97}} & \textbf{47.16} \\
\gptfouro & 42.82 & \textbf{42.98} & \textbf{41.77} & \textbf{\underline{45.89}} & \textbf{44.69} & 46.15 & 46.19 & 46.18 & \underline{47.93} & 46.81 \\
   \midrule
  \midrule
 \multicolumn{11}{l}{\quad \quad \quad \quad \quad \quad \quad \quad \textit{To English translation, incl. \textsc{De-En}, \textsc{Zh-En}}} \\ 
 \nllbsmall  & 29.52 &  &  &  &  & 34.35 &  &  &  & \\
\nllbmedium & 33.11 &  &  &  &  & 36.83 &  &  &  &  \\
\nllbbig & 33.82 &  &  &  &  & 37.89 &  &  &  & \\
\alma & 33.00 &  &  &  &  & 37.72 &  &  &  & \\
\qwen & 31.84 & 32.07 & 31.43 & \underline{35.65} & 34.61 & 38.19 & 38.39 & 36.72 & \underline{41.01} & 39.91 \\
\mistral & 27.26 & 29.73 & 27.55 & \underline{34.19} & 33.24 & 34.08 & 34.68 & 33.54 & \underline{38.39} & 36.94 \\
\gemma & 34.03 & 34.96 & 33.61 & \underline{38.98} & 38.74 & 38.65 & 38.86 & 38.35 & \underline{42.31} & 41.75 \\
\llamasmall & 30.71 & 33.47 & 30.96 & \underline{37.50} & 36.86 & 34.94 & 35.59 & 35.66 & \underline{39.06} & 39.03 \\
\llamalarge & \textbf{34.64} & \textbf{35.33} & \textbf{35.86} & \underline{\textbf{40.26}} & \textbf{38.94} & 38.09 & 38.66 & \textbf{39.40} & \textbf{42.39} & \underline{\textbf{42.42}} \\
\gptfouro & 34.57 & 34.47 & 34.15 & \underline{37.87} & 37.75 & \textbf{39.48} & \textbf{39.35} & 38.41 & \underline{42.04} & 41.36 \\
 \bottomrule                 
  \end{tabular}

 }
   \caption{CHRF++ scores on Proverb in Conversation Translation. \textbf{Bold} highlights the best score in each column. The better score among different prompts are \underline{underlined}.
   }
   \label{tab:chrf-bleu}
   \end{center}
\end{table*}

\subsection{Data Contamination Analysis}
\label{appx:data-contamination}
We measure the contamination rate of the LLM using a method similar to \citet{Liu2024AreMultilingualLLMs}. Specifically, for a given sentence and its preceding context $(s_{t}, s_{<t})$, we create an input prefix by truncating $s_{t}$ to a certain proportion of its words ($\tau$), denoting the prefix as $s_{t}^{<\tau}$ and the remaining part as the suffix $s_t^{\geq \tau}$. Two types of prompts are then created, one with context ($X_{c} = [s_{<t}; s_{t}^{<\tau}]$) and one without context ($X_{\emptyset} = s_{t}^{<\tau}$), where $[;]$ represents string concatenation.

We use the base version of the LLM\footnote{To reduce costs, we evaluated four LLMs at the 7-9B scale.} to complete the given prefix in both forms, resulting in hypotheses denoted as $\hat{Y}^{c}$ (with context) and $\hat{Y}^{\emptyset}$ (without context). During the generation, the greedy search strategy (temperature set as 0) is used to ensure the reproduction of the result.

Finally, we compute the length of the longest common subsequence (LCS) between the model’s predictions (with and without context) and the reference suffix, denoted as $|\text{LCS}(\hat{Y}^{c}, s_t^{\geq \tau})|$ and $|\text{LCS}(\hat{Y}^{\emptyset}, s_t^{\geq \tau})|$. The contamination ratio $\gamma$ is then estimated as:

\begin{align} \gamma = \max(0, \frac{|\text{LCS}(\hat{Y}^{c}, s_t^{\geq \tau})| - |\text{LCS}(\hat{Y}^{\emptyset}, s_t^{\geq \tau})|}{|s_t^{\geq \tau}|}) \label{eq:gamma} 
\end{align}

A higher value of $\gamma$ indicates a greater likelihood that the model has been affected by data contamination for a given sample. The reasoning behind this is as follows:
\emph{(i)} A large $\gamma$ occurs only when the first term ($|\text{LCS}(\hat{Y}^{c}, s_t^{\geq \tau})|$) is significantly larger than the second term ($|\text{LCS}(\hat{Y}^{\emptyset}, s_t^{\geq \tau})|$), indicating that the model can predict the suffix accurately when given context but struggles without it.
\emph{(ii)} When both terms are large, it likely means the sample is a proverb or well-known phrase, which the model can predict accurately even without context, resulting in a small $\gamma$.
\emph{(iii)} When both terms are small, it suggests that the model lacks sufficient knowledge to predict the suffix, with or without context.


\subsection{Prompt Template for Translation}
\begin{figure*}
    \centering
    \includegraphics[width=1.0\linewidth]{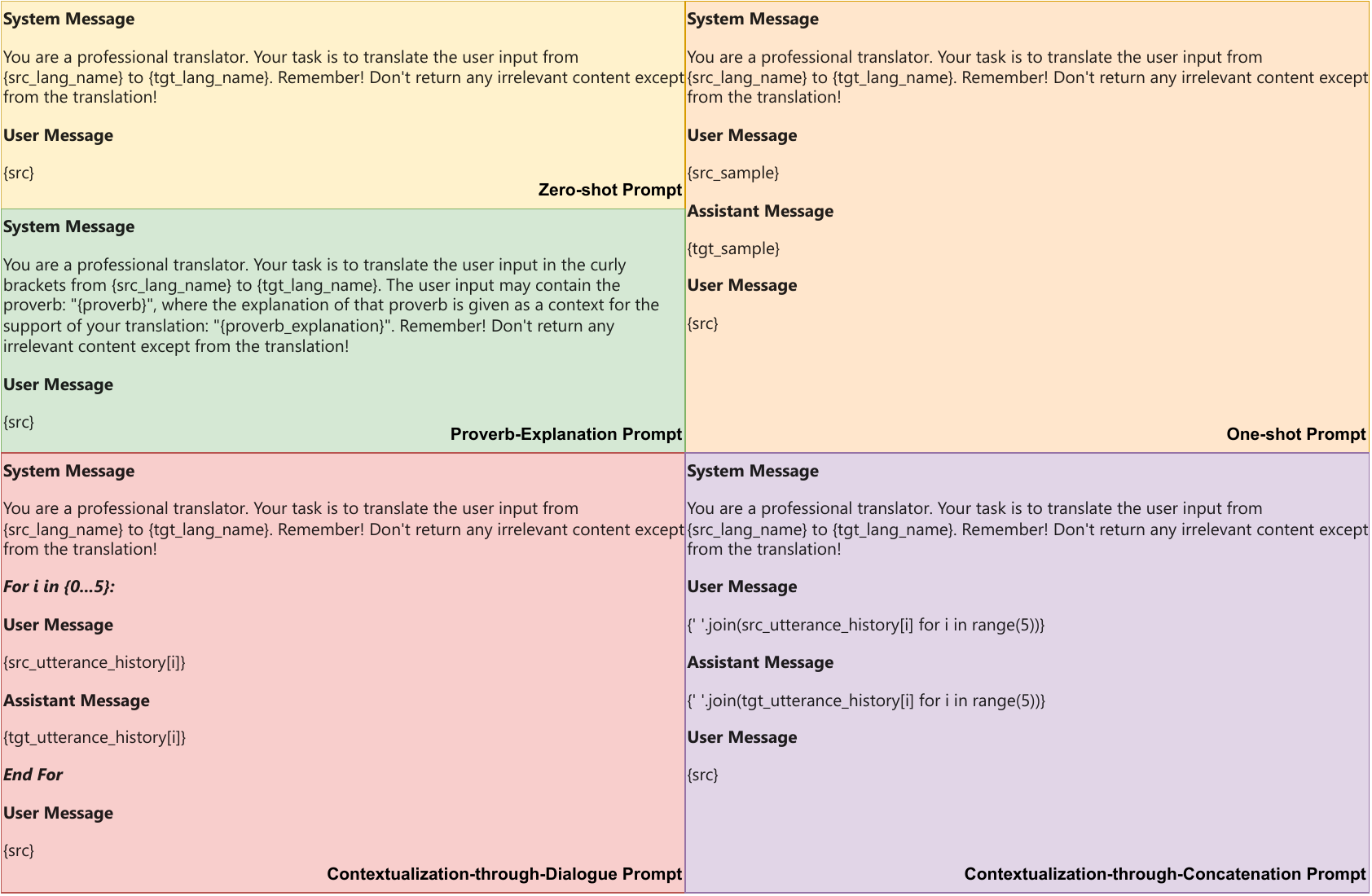}
    \caption{The prompt template of subtitle translation.}
    \label{fig:subtitle_trans_prompt_template}
\end{figure*}

\subsection{Prompt Template for LLM-based Evaluation}
\begin{figure*}
    \centering
    \includegraphics[width=0.8\linewidth]{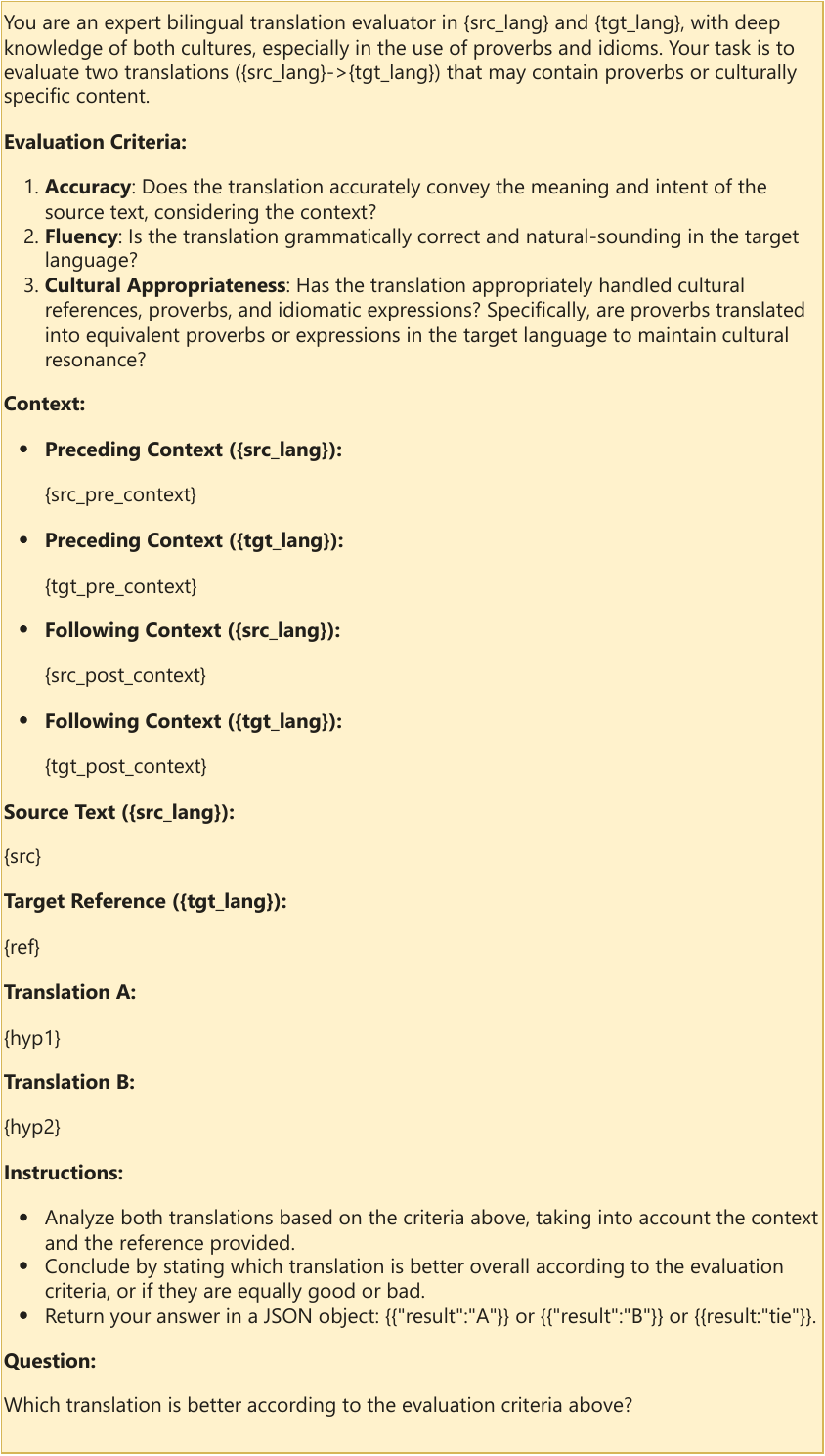}
    \caption{The prompt template of LLM-based evaluation.}
    \label{fig:llm_eval_prompt_template}
\end{figure*}

\end{document}